\documentclass[english]{article}
\usepackage{arxiv}
\usepackage[T1]{fontenc}
\usepackage[latin9]{inputenc}
\usepackage{color}
\usepackage{textcomp}
\usepackage{graphicx}

\makeatletter

\usepackage{cite}

\makeatother

\usepackage{babel}
\begin{document}
\title{A physics-guided neural network for flooding area detection using
SAR imagery and local river gauge observations}
\author{Monika Gierszewska, Tomasz Berezowski\thanks{The authors were with the faculty of Electronics, Telecommunications
and Informatics, Gda\'nsk University of Technology. Gabriela Narutowicza
11/12, 80-233  Gda\'nsk, Poland, e-mail: tomberez@eti.pg.edu.pl}}
\maketitle

\begin{abstract}
The flooding extent area in a river valley is related to river gauge
observations. The higher the water elevation, the larger the flooding
area. Due to synthetic aperture radar\textquoteright s (SAR) capabilities
to penetrate through clouds, radar images have been commonly used
to estimate flooding extent area with various methods, from simple
thresholding to deep learning models. In this study, we propose a
physics-guided neural network for flooding area detection. Our approach
takes as input data the Sentinel 1 time-series images and the water
elevations in the river assigned to each image. We apply the Pearson
correlation coefficient between the predicted sum of water extent
areas and the local water level observations of river water elevations
as the loss function. The effectiveness of our method is evaluated
in five different study areas by comparing the predicted water maps
with reference water maps obtained from digital terrain models and
optical satellite images. The highest Intersection over Union (IoU)
score achieved by our models was 0.89 for the water class and 0.96
for the non-water class. Additionally, we compared the results with
other unsupervised methods. The proposed neural network provided a
higher IoU than the other methods, especially for SAR images registered
during low water elevation in the river.
\end{abstract}

\section{Introduction }

Flooding is one of the most common natural phenomena occurring and
impacting people\textquoteright s lives and the environment around
the whole world. One important aspect of flood monitoring is estimating
its extent to calculate economic losses and investigate the effect
on the natural environment. Currently, the flood monitoring methods
include hydrological station monitoring, hydrodynamic modeling, and
remote sensing monitoring. The first method is characterized by high
accuracy; however, it is limited by the number of gauge stations,
and it monitors the situation only in a small area. Hydrodynamic modeling
is the most accurate method of estimating flooding extent, however,
it requires excessive study site parametrization along with the monitoring
data. The remote sensing methods are currently often applied due to
their increasing temporal resolution and ability to monitor floods
in large areas with sparse monitoring networks \cite{Xin2018,Raettich2020}.
Remote sensing data from both passive and active sensor types were
successfully applied in flood mapping, however, due to the frequent
cloud cover and rain during flood events, the number of available
cloudless optical images is strongly limited. These limitations do
not hold for the Synthetic Aperture Radar (SAR) data. Therefore, SAR
data is currently gaining more and more popularity if flood mapping
applications.

\subsection{Background and related work}

Contemporary methods of flood mapping using SAR images can be grouped
based on the mapping technique. An effective tool for delineating
the inundated areas is thresholding of SAR backscattering \cite{Yamadaa}.
The threshold can be determined in an automated way \cite{Long2014}
or by expert judgment \cite{Manjusree2012}. The computational efficiency
of this method makes it suitable for rapid mapping purposes; however,
its performance decreases in the case of large-area water bodies \cite{Tan},
complex topography, or dense vegetation occurring in the study area.

In contrast to pixel-based thresholding, image segmentation techniques
were proposed. They offer the advantage of using the spectral-spatial-related
characteristics of the parameters of the segments. Horrit et. al.
\cite{Horritt2001} used the statistical active contour method to
delineate a flood. However, this algorithm requires manual initialization
of the contour close to the edge of the object being segmented. Martinis
et. al. \cite{Martinis2009} proposed the split-based approach  what
is an automated method and offers some advantage over the thresholding. 

Another common technique is image change detection, usually performed
by comparing pre- or post-disaster reference data with flood imagery
\cite{Long2014,Bazi2005} by image difference, normalized difference,
image ratio, and log-ratio \cite{Manavalan2016,Chini2017}. Shen et
al. \cite{Shen2019} proposed an improved change detection (ICD) method
that uses multiple (\textasciitilde 5) pre-flood SAR images and a
multi-criteria approach, which leads to minimization of false positives
caused by water-like bodies and reduces the effect of speckles.

In general, the flood mapping approaches are based on a single or
dual image (pre- and post-event). Time-series data are less commonly
applied; however, utilizing this has been very useful in studies focusing
on flood monitoring and ecology \cite{Frappart2005,Ahmad2019,Schlaffer2018}
and hydrodynamic model calibration and validation \cite{Pulvirenti2014}.
Multi-temporal images also help reduce seasonal noise in the reference
image \cite{Schlaffer2015}. The pixel-wise methods using time-series
data are dependent on calculated features or manual threshold adjustment
and do not consider neighboring pixels \cite{Cian2018}. Compared
to them, deep learning methods can easily learn patterns in spatial
and temporal contexts without the process of feature selection, based
only on SAR polarization images.

Recently, most of the proposed flood mapping methods have been based
on applying deep learning models performing semantic segmentation,
which solved image segmentation at the semantic level and classified
images at the pixel level. The growth of its usage is caused by the
introduction of a convolutional neural network (CNN) that has shown
promising results in SAR data classification \cite{Ball2017,Ma2019}.
In the literature, there are many studies regarding flood or water
mapping methods that apply different deep learning architectures.
For example, Rizwan et al. \cite{Sadiq2022} investigated the performance
of U-Net and FCN. As shown in their study, deep learning models performed
better than Otsu due to the generalization capabilities of deep learning-based
segmentation models. Another study investigated the accuracy of FWENet
model water body extraction compared to the UNet, Deeplab v3, and
UNet++ models \cite{Wang2022}. FWENet performed the best due to applied
dilated convolution. Another approach proposed a multimodal OmbriaNet
architecture, which employed satellite data from various sensor types
and timestamps (pre- and post-event images) \cite{Drakonakis2022}.

\subsection{Aim and motivation}

Several studies showed the benefit of connecting the data from hydrological
and satellite monitoring. The river gauge observations, such as water
elevation and discharge, are related to the flooding extent area in
a river valley \cite{Arnesen2013}. The higher the observed water
elevation or discharge, the larger the flooding area. For example,
researchers used the flooding extent maps to derive time series of
river discharge estimates in several British Columbia (Canada) \cite{Smith1996}
and South Korea \cite{Ahmad2019} rivers. Another study related the
river gauge observation in the Congo River (Africa) with SAR backscatter
using satellite altimeter data instead of measured water elevations
\cite{Kim2017}. Yet another study demonstrated that the relation
between inundation extent and river gauge observations can be conversely
used \cite{Berezowski2020,Berezowski2024}.  The results showed that
the relation is close to linear, and hydrological observation can
be applied to SAR-derived flooding extent mapping.

The presented supervised deep learning approaches provide highly accurate
results. However, they require data annotation, for which acquisition
is a time-consuming process that demands expert knowledge. To overcome
this limitation, some unsupervised deep learning methods were proposed.
One example of an unsupervised method is a model based on a spatiotemporal
variational autoencoder that was trained with reconstruction and contrastive
learning techniques \cite{Yadav2024}. Apart from their successes
in flood detection using SAR imagery, this approach is weakly supervised
and still relies on training/labeled samples. Another approach combined
super-pixel and CNN to map floods using SAR imagery and routinely
available ancillary datasets \cite{Jiang2021}. This method was applied
only for VV-polarized SAR data that have an influence on the discrimination
between flooded and non-flooded areas. Similarly, for single-polarized
SAR data, another study presented a fully automated approach, including
the process of obtaining training labels from auxiliary data such
as DTM and optical images, and the supervised deep learning model
\cite{Nemni2020}. Despite the described approaches, there is still
a need to develop unsupervised machine-learning techniques for water
extent mapping on time-series satellite data \textcolor{black}{\cite{Ma2019,Parikh2019}.}

Most of the presented studies performed the detection task of water
extent during flood events. The supervised deep learning methods were
usually trained with satellite images showing flood areas. However,
the studies did not show the efficiency of water area mapping of the
proposed methods for images acquired during non-flood conditions.
Some other research showed the algorithms for river channel detection\textcolor{black}{{}
\cite{Klemenjak2012,Han2023,Gasnier2021}. These methods e}xtracted
rivers of SAR images registered for low water elevation accurately.
However, there are no experimental results of mapping accuracy in
the case of inundated areas occurring in the river valley. Hence,
it will be valuable to perform a method that will be efficient in
water area detection regardless of the water elevation in the river.

One of the new trends is to integrate domain knowledge and achieve
physical consistency by teaching machine learning models about the
governing physical rules of the Earth system. This approach, known
as physics-guided neural networks (PGNN), combines deep learning with
physics-based models. PGNNs can be created in two different ways:
by combining the training data with the results of the physical model
or by adding to the cost function the physical formula that ensures
that the predictions are physically consistent. An example of a PGNN
approach is a time-varying lake temperatures model \cite{Daw2022}.
In particular, that study integrated the physical relationships between
the temperature, density, and depth of water into the PGNN. Another
example of a PGNN integrated augmented satellite image samples with
pixel-level physical parameters derived from Landsat 8 images for
water leak detection in canals \cite{Chen2020}. Yet another study
built a CNN model architecture inspired by polarimetric features for
SAR image classification \cite{Huang2022}. The PGNN approach outperformed
the deep learning models regarding prediction accuracy, especially
with a limited input dataset \cite{Read2019}.

Our study aimed to create an unsupervised method for water extent
mapping that would help to identify river and non-river flooding in
locations with sparse observation networks, various climate conditions,
and different land-cover types. We use the Sentinel 1 time series
SAR images in VV and VH polarization as input data and water elevation
data from river gauges as labels. Based on the described studies \cite{Arnesen2013,Smith1996,Ahmad2019,Kim2017,Berezowski2020},
we can assume that there is a monotonically non-decreasing relation
between a local river water elevation and the flooded area, and it
is close to linear. This dependence can be quantified using the Pearson
correlation coefficient. We utilize the dataset to train the physics-guided
neural network for flooding area detection (FPGNN). By applying the
correlation coefficient between local water elevation and water area
on SAR images as a loss function in the neural network, we show that
the model predictions are consistent with the assumption. Next, we
quantify the model performance with ground truth data from DTM and
optical images. In the final step, we compare our method to other
automated approaches and discuss the results depending on the study
areas\textquoteright{} characteristics.

\section{Study areas and datasets}

This section outlines the study areas, the acquisition of the input
data (hydrological and SAR data), the preprocessing steps applied
to the SAR data, and the reference data used in the validation process. 

\subsection{Study sites}

To show the performance of our method, we chose five study cases that
varied by climate, river width, and land use. The first case shows
the Parana Madeirinha River located in Brazil, in the period January
2019 to December 2022 (119 images). The climate is tropical monsoon,
and the land use is forest, barren, and urban. The width of the riverbed
is approximately 280 m, and the maximum difference in water level
is 12.59 m. The chosen water station is located in Autazes city (59.1$^{\circ}$
W, 3.6$^{\circ}$S). The second case is the Mekong River located in Cambodia,
in the period March 2017 to December 2018 (130 images). The climate
is tropical savanna, and the land use is mostly agricultural land
with small villages along the river. The width of the riverbed is
approximately 500 m and the maximum difference in water level is 8
m. The hydrological observation was obtained from a gauge located
in Prek Kdam city (104.8$^{\circ}$E, 11.8$^{\circ}$N). The third case contains four
floods that occurred on the Po River in Italy, in the period November
2014 to October 2020 (120 images). The climate is humid subtropical,
and the land is covered by agricultural fields and forests. The river
is about 200 m wide. The water gauge station is located under the
bridge near Valenza city (8.6$^{\circ}$E, 45$^{\circ}$N) and the maximum difference
in water level is 5.70 m. The final two study cases are located in
England (Derwent River -- 144 images) and Poland (Warta River --
145 images) with oceanic and humid continental climates, respectively.
Floods on both rivers occur usually during the winter and spring seasons.
Most of the areas are agricultural and there are flood embankments
along both rivers. The main difference between these study cases is
in the width of the rivers -- 20 m in England and 100 m in Poland.
The water elevation data used were measured at the Bubwith (-0.9$^{\circ}$W,
53.8$^{\circ}$N) and \'Swierkocin (15$^{\circ}$E, 52.6$^{\circ}$N) stations, respectively.

\subsection{SAR Data and Processing}

The Sentinel-1 radar mission is composed of a constellation of two
satellites, Sentinel-1A (launched on 3 April 2014) and Sentinel-1B
(launched on 25 April 2016). The revisit time depends on the region;
the longest is at the equator, at 6 days. Sentinel-1 sensors transmit
C-band waves in dual-polarization mode and the data are freely available.

In this study, we used the Level-1 Ground-Range Detected High-Res
Dual-Pol (GRD-HD) Sentinel 1 products in the interferometric wide
swath (IW). We searched for the Sentinel 1 data available for different
time ranges for each test site on the Alaska Satellite Facility (ASF).
We used Radiometrically Terrain Corrected (RTC) Sentinel-1 products
preprocessed on-demand by the ASF Hybrid Pluggable Processing Pipeline
(HyP3) \cite{Hogenson2016}. Processing with this software consisted
of five steps: co-registration to the DEM file, radiometric calibration
to the backscatter coefficient ($\gamma{}_{0}$) {[}dB{]}, speckle
filtering using the Enhanced Lee filter in a 7 \texttimes{} 7 pixel
kernel, range-Doppler terrain correction with projection to the UTM
coordinate system with proper zone, and finally cropping the images
to the proper bounding box of the study area. Every downloaded image
contained two bands: Gamma0\_VV and Gamma0\_VH. The spatial resolution
of the images was 10 m and the image size was different for each study
area. 

\subsection{Hydrological Data}

We obtained daily hydrological data from the gauge station located
in each of the study areas \cite{Brazil,Italy,Mekong,Poland_wl,UK_wl}.
The water elevations were directly measured on the gauge. We recalculated
the water elevations in the validation step by adding a height above
sea level zero of the water gauge to the measured elevations. 

\subsection{Reference Data}

\subsubsection*{1) Elevation Data}

We used the 30 m Copernicus DEM for the Cambodia and Italy sites.
For the relevant area of Brazil, the Copernicus DEM had inaccurate
elevation due to registration during high water elevation in the river.
Therefore, we chose elevation data provided by the Shuttle Radar Topography
Mission (SRTM) for this site. For areas in Poland and England, the
digital elevation models produced by government agencies \cite{Uk_dtm,Poland_dtm}
were used, because they had higher spatial resolution than the Copernicus
and SRTM DEMs.

\subsubsection*{2) Optical Data}

As the source of optical data, we chose the Landsat 8 (L8) Operational
Land Imager (OLI) and the Sentinel-2 (S2) MultiSpectral Instrument
(MSI). The images were acquired by Sentinel Hub API. We filtered the
images by collection (Sentinel 2 L2A, Landsat 8 L2A), acquisition
time (the same time range as for the Sentinel 1 data), percentage
of cloud cover, and area of interest. Next, we calculated the Modified
Normalized Difference Water Index (MNDWI) \cite{Xu2006} using bands
3 and 6 for the L8 data and bands 3 and 11 for the S2 data. Then,
we resampled the optical data to 10 m resolution and clipped them
to the same area as the SAR data. Additionally, we manually selected
the clouds and cloud shadows pixels on the optical images.

\section{Methodology}

\begin{figure*}
\begin{centering}
\includegraphics[width=1\textwidth]{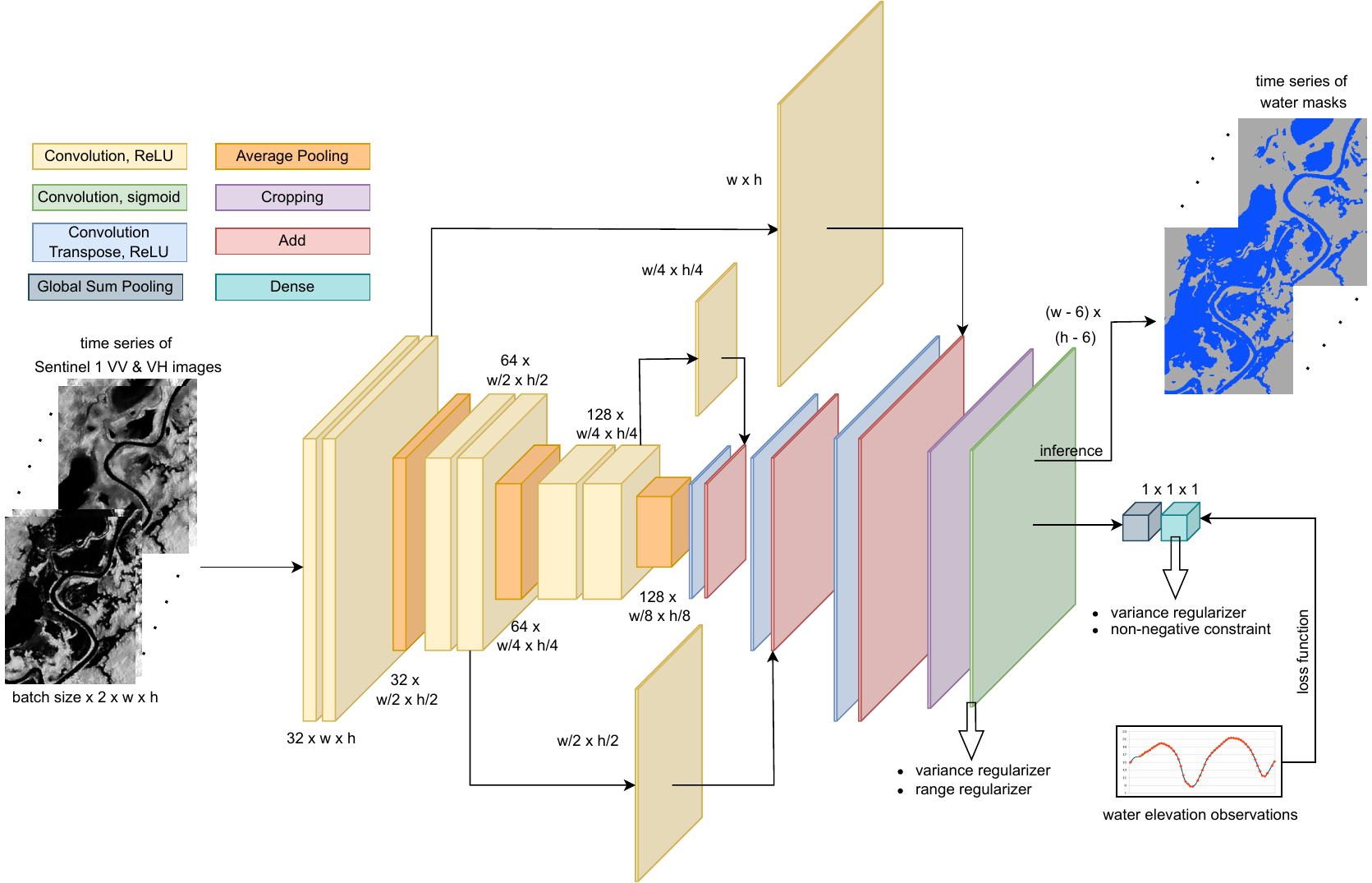}
\par\end{centering}
\caption{\label{Fig 1}Architecture of the FPGNN model.}
\end{figure*}

In this section, we introduce our proposed approach to water extent
mapping by describing the architecture of the proposed neural network
and validation steps. The PGNN model architecture is outlined in Fig.
\ref{Fig 1}.

\subsection{Network architecture}

Our method for mapping the extent of the water is based on the FCN-8
fully convolutional network \cite{Shelhamer_2017} architecture, which
was one of the first CNN architectures for semantic segmentation.
We built our algorithm by reducing the number of layers in the FCN
architecture and adding regression layers. In our FPGNN architecture,
we can highlight three modules. The first module, named the feature-extraction
module, produces the features obtained from input data by convolutional
function and image size reduction. This part of the network consists
of six 3x3 convolutional layers with the ReLU activation function
and three 2x2 average pooling layers with stride 2. 

Next, as in a FCN, the output of the first module is used as input
to the deconvolutional layers, which restore the small size feature
maps to their original size. We also inserted an \textquotedblleft add\textquotedblright{}
layer that sums the deconvolved output and the convolved output (from
the feature extraction module in which number of channels is earlier
averaged to one channel using a 1x1 convolutional layer). The upsampling
part contained three 4x4 deconvolutional layers with ReLU activation
and stride 2, three \textquotedblleft add\textquotedblright{} layers
and one 1x1 convolutional layer to integrate the output of the previous
layer into a consistent features set . The output of the convolution
layers was padded with zeros to ensure that the size of the last two-dimensional
layer in our network is the same as the input size. The purpose of
the last convolutional layer was to provide the water body segmentation
map, \textbf{hereinafter named the water mask}\emph{.} Therefore,
in contrast to the softmax function that was applied in the FCN, the
activation function used here was sigmoid, which is commonly used
in binary segmentation task because the output of this function will
range between 0 and 1. 

The last part of the architecture is the regression module. It includes
a global sum pooling layer that sums all pixels from the water mask
to one value. The calculated sum is the input to the last fully connected
layer without an activation function to get the final output (water
extent area), which is evaluated against the ground truth local water
elevation by a correlation loss function (eq. \ref{eq:1}). 

\subsection{Training Algorithm}

Our dataset for each test site contained time-series of dual-pol (VV+VH)
radar images with sizes depending on the study sites, and the corresponding
time-series of local water elevations. This dataset was divided into
training and testing sets with an 80\% to 20\% ratio. The radar data
presented the riverbeds at different stages: low, average, or high.
Hence, we used stratified sampling to ensure that samples for various
water elevations were present in both the training and test sets. 

In model training, we used the Adam optimization algorithm with a
learning rate of 0.01. The batch size used was equal to half the number
of samples available in the training set. As a loss function, we adopted
(eq. \ref{eq:1}) the Pearson correlation coefficient (PCC) between
the predicted sum of the water mask area, i.e., the output of the
global sum pooling layer ($\hat{y}$), and the local water level observations
($y$) of river water elevations. 
\begin{equation}
Loss=1-\frac{{\displaystyle \sum_{i=1}^{n}(\hat{y}_{i}-\bar{\hat{y}})(y_{i}-\bar{y}})}{\sqrt{{\displaystyle \sum_{i=1}^{n}(\hat{y}_{i}-\bar{\hat{y}})^{2}}}\sqrt{{\displaystyle \sum_{i=1}^{n}(y_{i}-\bar{y}})^{2}}}\label{eq:1}
\end{equation}
where $n$ is the set size,$\bar{\hat{y}}$ is the mean value of $\hat{y}$,
and $\bar{y}$ is the mean value of $y$. The values of PCC are always
between -1 (negative correlation) and 1 (positive correlation). A
value of 0 indicates that there is no correlation between the two
variables.

As the model is trained without labeled data, we added a modification
to the network training to help it make predications. In the last
model layer, i.e., the global sum pooling, we added a kernel constraint
that maintains the nonnegative weights. This ensures that the model
set higher values for each water pixel than for each non-water pixel
in the output of the upsampling module. Effectively, the water pixels
in the water body segmentation map are forced to converge to one and
non-water pixels converge to zero. Our model sometimes tended to provide
the wrong prediction despite the obtained high correlation coefficient
of the training and validation sets. For example, the difference in
values of water and non-water pixels in the output result of the upsampling
module were very low, or the model did not retrieve all pixel values
after the deconvolution layers in the output of the upsampling module.
To avoid these, we created a custom activity regularizer that applies
a penalty to the layer\textquoteright s output. The penalty is calculated
for the batch and divided by the batch size. In the fully connected
layer, we applied the activity regularizer calculated by $\frac{1}{0.01*\sigma^{2}}$
, where $\sigma^{2}$ is the variance of the predicted output. Another
custom activity regularizer (eq. \ref{eq:2}) was added to the convolutional
layer with the sigmoid activation function, i.e., the water mask,
as: 
\begin{equation}
reg=10*\frac{1}{max(x)-min(x)}+100*\sigma_{clip}^{2}\label{eq:2}
\end{equation}
where $x$ is the layer output and $\sigma_{clip}^{2}$ is the variance
calculated for the subset of the layer output (the matrix of 6x6 pixel
size belonging to the non-water class on each dataset sample). We
applied the early stopping method in the training process. This method
stops training when a monitored metric (validation loss) has stopped
improving. We also added a new condition that checks the difference
in the maximum and minimum values in the output of the upsampling
module. The training is stopped if the difference is higher than 0.9
and the validation loss does not improve. If the difference is lower,
the training is continued. This condition improves the result of the
output of the upsampling module that produces the water body segmentation
map. We trained the network for each study area separately, producing
five trained models. Based on each trained model, we created an intermediate
model for inference with the last convolutional layer with a sigmoid
function as the output layer. Due to that, the inference outputs were
images with sizes the same as the input radar images representing
a soft classification of water extent with pixel values between 0
and 1. Hard classification of water extent was obtained using thresholding.
We tested the effects for various thresholds in range from 0.1 to
0.55, where the pixels with a value above or equal to the threshold
were labeled as water class and pixels with values below the threshold
were classified as non-water class.

\subsection{Validation Process}

\begin{figure*}
\begin{centering}
\includegraphics[width=1\textwidth]{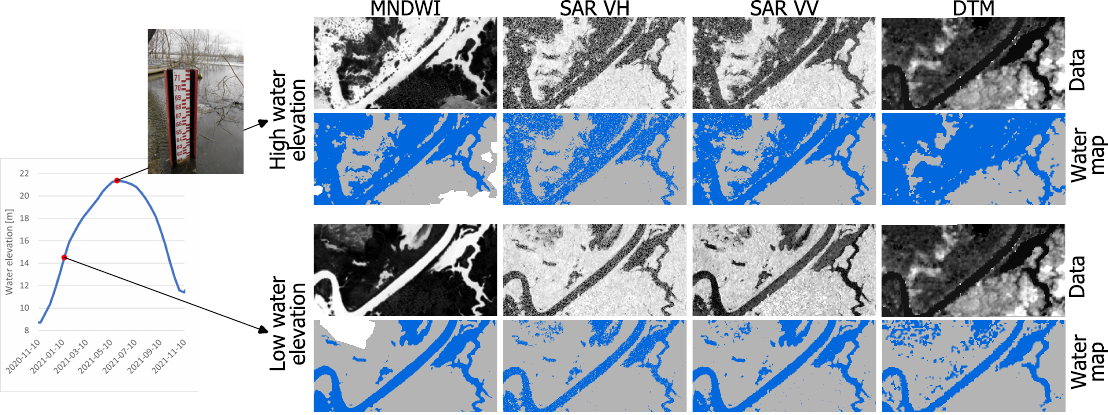}
\par\end{centering}
\caption{\label{Fig 2} Visual comparison of water masks obtained by thresholding
from various spatial data and the effect of water elevation in the
river on the water extent area.}
\end{figure*}

\subsubsection*{1) Validation using DTM-based water mask}

The water mask for the DTMs was extracted by labeling as water the
pixels with values lower than the water elevations recalculated to
zero-gauge level (Fig. \ref{Fig 2}). The time-series-masked DTMs
were provided for the same days as the SAR images, except for days
when the measured water elevation was lower than the elevation in
the river pixels present in a DTM. Effectively, a binary water map
for the same extent as the SAR images was produced. However, this
method was only valid for rivers with a gentle bed slope. In these
cases, the difference between the extrapolated water extent assuming
no riverbed slope and the true water extent was smaller than one SAR
pixel. For the rivers with steep riverbed slopes, which in our case
are located in Poland, England and Italy, we limited the DTM-based
validation to the extent near the water gauge.

\subsubsection*{2) Validation using MNDWI water mask}

To produce the water mask for the optical data, we used the MNDWI
images. In the MNDWI data, pixels with a value larger than zero are
classified as water (Fig. \ref{Fig 2}). However, the appropriate
threshold depends on the sensor, imaged region, and acquisition time
\cite{Du2016}. Hence, we calculated the threshold for each MNDWI
image using the Otsu thresholding method. The Otsu method automatically
determines the global threshold to separate pixels into two classes
by minimizing the variance between classes \cite{Otsu1979}. In the
cases where the image histogram was not bimodal, we performed manual
thresholding. Subsequently, we applied the clouds mask and cloud shadows
mask to the resulting maps. 

\subsubsection*{3) Validation metric}

The ground truth and predicted water cover maps were compared using
the Jaccard index, also named Intersection over Union (IoU, eq. \ref{eq:3}). It measures
the similarity between two binary images. 
\begin{equation}
IoU_{n}=\frac{A_{P,R}}{A_{P}+A_{R}-A_{P,R}}\label{eq:3}
\end{equation}
where $A_{P}$ is the area of $n$ class on the predicted mask, $A_{R}$
is the area of the $n$ class on the reference mask, and $A_{P,R}$
is the area of intersection of the $n$ class on both masks. In our
case, the $n$ class was either water or non-water. The values of
IoU range from 0 to 1, where zero means that the masks do not intersect,
and one means that the masks completely overlap. 

\subsubsection*{4) Benchmarking methods}

To investigate the effectiveness of our algorithm, we compared it
with other fully automatic methods, such as Otsu thresholding, Chan-Vese
algorithm, gaussian mixture model (GMM), and spectral clustering (SC).
In each of these methods, we applied SAR images converted to dB. In
the Otsu method, we calculated the Otsu threshold and used it to create
the water masks. The Chan-Vese segmentation algorithm is a technique
for image segmentation based on the level set method, particularly
effective for segmenting objects with blurry or poorly defined boundaries.
It is a region-based segmentation model that works by minimizing an
energy functional to detect the boundaries of objects within an im\textcolor{black}{age
\cite{Chan1999}. The nex}t method we applied was the Gaussian mixture
model. It is an unsupervised clustering algorithm that assumes input
data conforms to a normal distribution. The GMM models the mixture
distribution of input data as follows \cite{Bishop2006} :
\begin{equation}
p(x)=\mathop{\mathop{\sum_{c=1}^{M}\pi_{c}=1.N(x|\varepsilon_{c},\Sigma_{c})(0\leq\pi_{c}\leq1)}}\label{eq:3-1}
\end{equation}
where $x$ refers to D-dimensional observation data; $M$ is the total
number of mixture components; $\pi_{c}$ is the weight factor of c-th
mixture component and satisfies $\sum_{c=1}^{M}\pi_{c}=1.N(x|\varepsilon_{c},\Sigma_{c})$
represents the estimated Gaussian density of c-th mixture component,
and the parameters $\varepsilon_{c}$ and $\Sigma_{c}$ are its mean
vector and covariance matrix, respectively. The last method was spectral
clustering. Unlike traditional clustering algorithms like K-means,
which rely on distance-based criteria, spectral clustering works by
utilizing the eigenvalues and eigenvectors of a similarity graph to
cluster the points. It is very useful when the structure of the individual
clusters is highly no\textcolor{black}{n-convex \cite{Luxburg2007}. }

\section{Results and Discussion}

This section includes the analysis of results and water extent maps
obtained with our methodology. It also contains an evaluation of the
efficiency of our approach compared to other methods.

\subsection{Results of validation process}

\begin{figure*}
\begin{centering}
\includegraphics[width=1\textwidth]{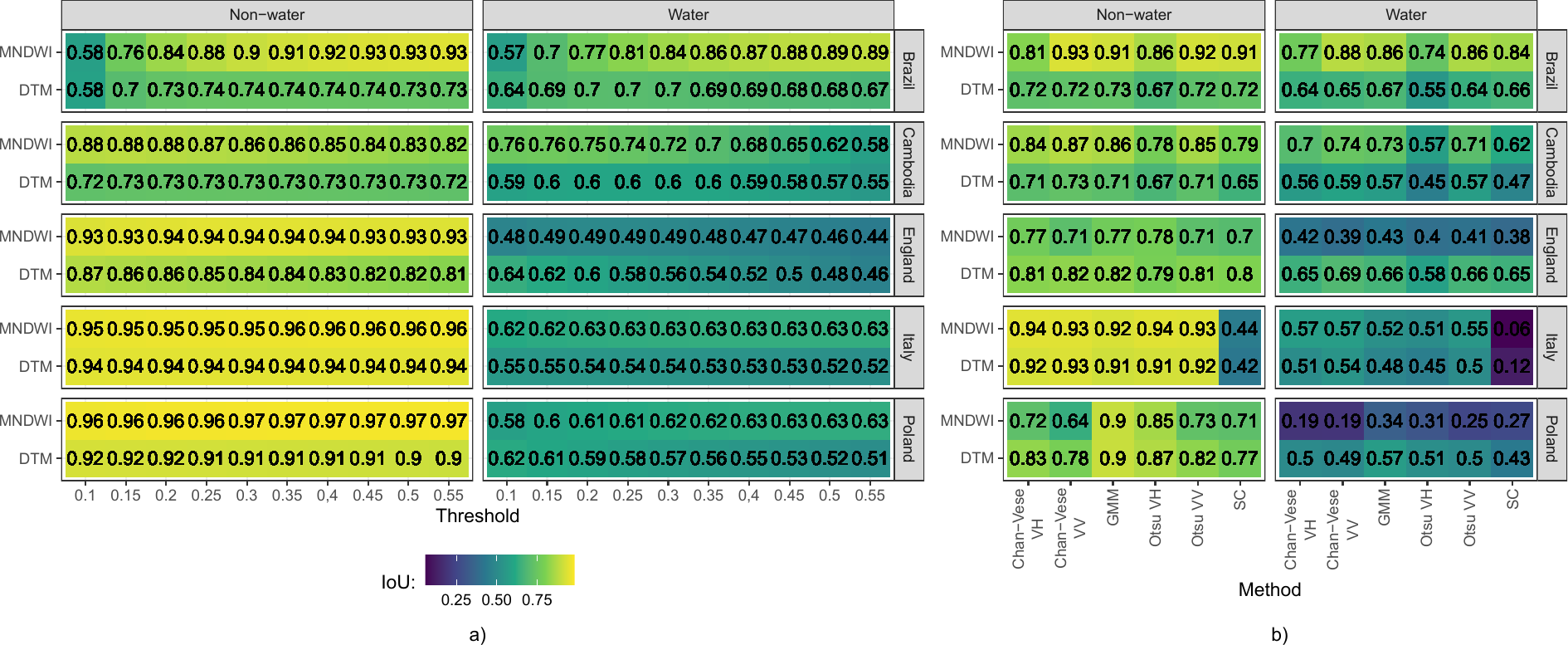}
\par\end{centering}
\caption{\label{Fig 3} Average IoU values of water  and non-water masks produced
by FPGNN models (a) or benchmarking methods (b). Panels are varied
by the study sites (columns) and class (rows). Within each panel,
the IoU value is shown per threshold used in hard classification or
benchmarking method (panel columns) and reference water mask (panel
rows). }
\end{figure*}

The threshold used to create the binary masks from the FPGNN results
ranged between 0.1 and 0.55 with a 0.05 step. Pixels with values below
the threshold were classified as non-water, and pixels above or equal
to the threshold were labeled as water class. The IoU values were
obtained for each SAR image with accompanying DTM-based or MNDWI water
mask and averaged in the study period. Depending on the threshold
applied in the validation process, the IoU values obtained were different
(Fig. \ref{Fig 3}a). For areas located in Poland and England, the
average IoU of the water class in validation with the DTM decreased
as the threshold increased. In contrast, in Brazil\textquoteright s
example, the IoU was the highest for the 0.35 threshold. The IoU values
did not change significantly depended on the threshold for the Cambodia
and Italy areas. In the case of the MNDWI validation, the IoU for
the water class was greater for a smaller threshold in the Cambodia
and England areas. The threshold did not have a significant impact
on the results for the Italy region, where the IoU values differ by
only about 0.01 regarding on threshold. The increasing threshold increased
the IoU values considerably in the Brazil model and slightly in the
Poland model. In terms of the ground truth data used in the validation,
the average IoU for both classes were usually higher for comparison
results with the MNDWI images than with the DTM (except for the water
class in the England study area). Significant differences ($p<2.2*10^{-16}$)
between the IoU for the MNDWI and DTM occurred in the Brazil and Cambodia
study areas, where the elevation data applied had a spatial resolution
of 30 m.

The main advantages of DTM water masks are their availability with
the same frequency as hydrological observation and the detection of
water beneath vegetation. However, the accuracy of the obtained water
extent depends strongly on the spatial resolution, the occurrences
of areas located below the water elevation outside the river valley,
and the slope of the riverbed. Another disadvantage is that this method
does not take into account water areas resulting from other sources
than high-water elevation in the river, such as heavy rainfall. Therefore,
water masks from the DTM are accurate only for the area near the water
gauge station and using high-quality elevation data. With the method
of thresholding the MNDWI images, one of the most important limitations
is the time resolution. The availability of optical images is strongly
limited by clouds, which often occur during flood events. Additionally,
the acquisition day of optical and radar images should be the same
due to the fact that flooding is a dynamic phenomenon (in our study,
we accepted a max +/- 4 days difference if no radar image was acquired
during the flood). Also, monitoring the presence of water under vegetation
is limited with optical satellite images, like similar with SAR images
registered in the C band, which is characterized by reduced canopy-penetrating
ability \cite{Gierszewska2022}. In our study, the water masks from
the MNDWI data overlap more with the water masks from the SAR images
than from the DTMs, which confirmed the results from the validation
process (Fig. \ref{Fig 3}a).

The overall average IoU of FPGNN for MNDWI validation ranged between
0.57 (Brazil model with a threshold of 0.1) and 0.91 (Brazil model
with a threshold of 0.5 and 0.55). For the average IoU calculated
per class, the water class had the highest average IoU (0.89) for
the Brazil model with a threshold of 0.5 and 0.55 in validation with
the MNDWI (Fig. \ref{Fig 3}a). The lowest average IoU for the water
class (0.44) was for the England area, with a threshold of 0.55. The
non-water class was classified more accurately than the water class.
The highest IoU for this class was 0.95--0.97 for comparison of the
model\textquoteright s Poland and Italy results (threshold 0.1--0.55)
with the MNDWI, and the lowest was 0.58 in MNDWI validation of the
Brazil model outputs thresholded at a value of 0.1.

The highest IoU (0.89) of the Otsu method (Fig. \ref{Fig 3}b) obtained
in MNDWI validation was for the VV data of the region in Brazil. However,
the lowest IoU (0.49) was obtained for the VV images from Poland in
comparison with the MNDWI water masks. Regarding the IoU per class,
the highest IoU of the water class was for the Brazil model performed
with VV data (0.86) and the lowest was for the Poland area, also with
VV images (0.25). The non-water class provided the highest IoU for
the VH Italy data (0.94), and the lowest for the England VV data (0.71).

In case of Chan-Vese segmentation, the highest IoU (0.90) for MNDWI
validation was for the VV data of the region in Brazil. However, the
lowest IoU (0.42) was obtained for VV images from Poland in comparison
to the MNDWI water masks. In case of IoU per class, the highest IoU
of the water class was in the Brazilian model performed with VV data
(0.88) and the lowest for the Poland area also with VH and VV images
(0.19). The non-water class provided the highest IoU of the VH Italy
data (0.94) and the lowest for the Poland VV data (0.64). 

The Gaussian mixture model provided the highest IoU (0.88) for MNDWI
validation for the region in Brazil. However, the lowest IoU (0.6)
compared to the MNDWI water masks was obtained for the Poland area.
In the case of IoU per class, the highest IoU of the water class was
in the Brazil model (0.86) and the lowest for the area of Poland (0.34).
The non-water class provided the highest IoU for Italy data (0.92)
and the lowest for England (0.77). 

In the case of the spectral clustering method, the highest IoU (0.88)
for MNDWI validation was the region in Brazil. However, the lowest
IoU (0.25) in comparison with MNDWI water masks was obtained for the
area of Italy. In the case of IoU per class, the highest IoU of the
water class was in the Brazil model (0.84) and the lowest for the
Italy area (0.06). The non-water class provided the highest IoU for
Brazil data (0.91) and the lowest for Italy (0.44). 

From benchmarking methods, the highest IoU values for MNDWI validation
were obtained from the Chan-Vese method applied to VV data (Brazil,
Cambodia, and Italy study areas) and the Gaussian mixture model (England
and Poland study areas). The evaluation of the water masks with MNDWI
images showed that the FPGNN outputs, which obtained the highest average
IoU from tested thresholds (0.5 for Brazil, 0.2 for Cambodia, 0.15
for England, 0.3 for Italy, 0.2 for Poland), provided equal IoU values
to best benchmarking methods for the Brazil and Cambodia data. The
higher IoU than other unsupervised methods, the FPGNN obtained for
Italy (0.04), England (0.11) and Poland (0.16) areas. 

For the DTM validation, the lowest average IoU obtained by FPGNN was
0.61 for Brazil, with a threshold of 0.1 and the highest was 0.77
for Poland, with a threshold of 0.1. The average IoU values calculated
per class were higher for the non-water class, except for the Brazil
site. The water class had the highest average IoU (0.7) for the Brazil
model, with a threshold of 0.2--0.35 (Fig. \ref{Fig 3}a). The lowest
average IoU (0.46) was in comparison with the DTM for the England
area with a threshold of 0.55. For the non-water class, the highest
IoU was 0.94 for the Italy results (threshold 0.1--0.55), and the
lowest was 0.58 for the Brazil model outputs thresholded at a value
of 0.1.

With the Otsu algorithm, the highest average IoU was 0.73 for the
England model trained with VV images, and the lowest (0.56) was obtained
for the VH Cambodia dataset. The water class had the highest IoU (0.66)
for the England VV images and the lowest (0.45) for the VV data from
Cambodia and Italy. The IoU for the non-water class in DTM validation
ranged between 0.67 (the results of the VH bands for Cambodia and
Brazil) and 0.92 (the results of the VH bands for Italy) (Fig.\ref{Fig 3}b).

In the case of the Chan-Vese method, the highest IoU (0.75) provided
for DTM validation was for the VV data of the region in England. However,
the lowest IoU (0.63) was obtained for VH images from Cambodia compared
with the DTM water masks. Regarding the IoU per class, the highest
IoU of the water class was for the Chan-Vese method in the England
model performed with VV data (0.69) and the lowest for the Poland
area with VV images (0.49). The non-water class provided the highest
IoU for VH Italy data (0.93) and the lowest for Cambodia VH data (0.71). 

For the Gaussian Mixture model, the highest IoU (0.74) provided for
DTM validation was for the region in England. However, the lowest
IoU (0.64) was obtained for Cambodia. In the case of IoU per class,
the highest IoU of the water class was for the Brazil study area (0.67)
and the lowest for the Italy area (0.48). The non-water class provided
the highest IoU for Italy data (0.91) and the lowest for Cambodia
data (0.71). 

For the spectral clustering method, the highest IoU (0.72) provided
for DTM validation was for the region in England. The lowest IoU (0.27)
was obtained for the Italy study area. Regarding the IoU per class,
the highest IoU of the water class was for the Brazil study area (0.66)
and the lowest for the Italy area (0.12). The non-water class provided
the highest IoU for England data (0.80) and the lowest for Italy data
(0.42). 

From benchmarking methods, the highest IoU values for DTM validation
were obtained from the Chan-Vese method applied to VV data (Cambodia,
England, and Italy study areas) and the Gaussian mixture model (Brazil
and Poland study areas). When comparing the FPGNN outputs, which obtained
the highest average IoU from tested thresholds, with the best unsupervised
method\textquoteright s results, the FPGNN method had similar or slightly
higher IoU for all study cases. In England, the water class was classified
better with Chan Vese segmentation with VV data, whereas the non-water
class obtained higher accuracy with FPGNN. However, the comparison
with the DTM is not reliable in the England area due to the fact that
when the elevation of the water in the river is low, the river patch
is not visible in the SAR images, whereas it is shown as water in
the DTM mask. Due to the numerous misclassified pixels on the benchmarking
method\textquoteright s results covered with water pixels on the DTM
water mask, the water mask from benchmarking methods had a higher
water class IoU than the FPGNN water mask, where no water pixels were
detected. 

\subsection{Impact of water elevation on effectiveness of FPGNN }

\begin{figure*}
\begin{centering}
\includegraphics[width=0.75\textwidth]{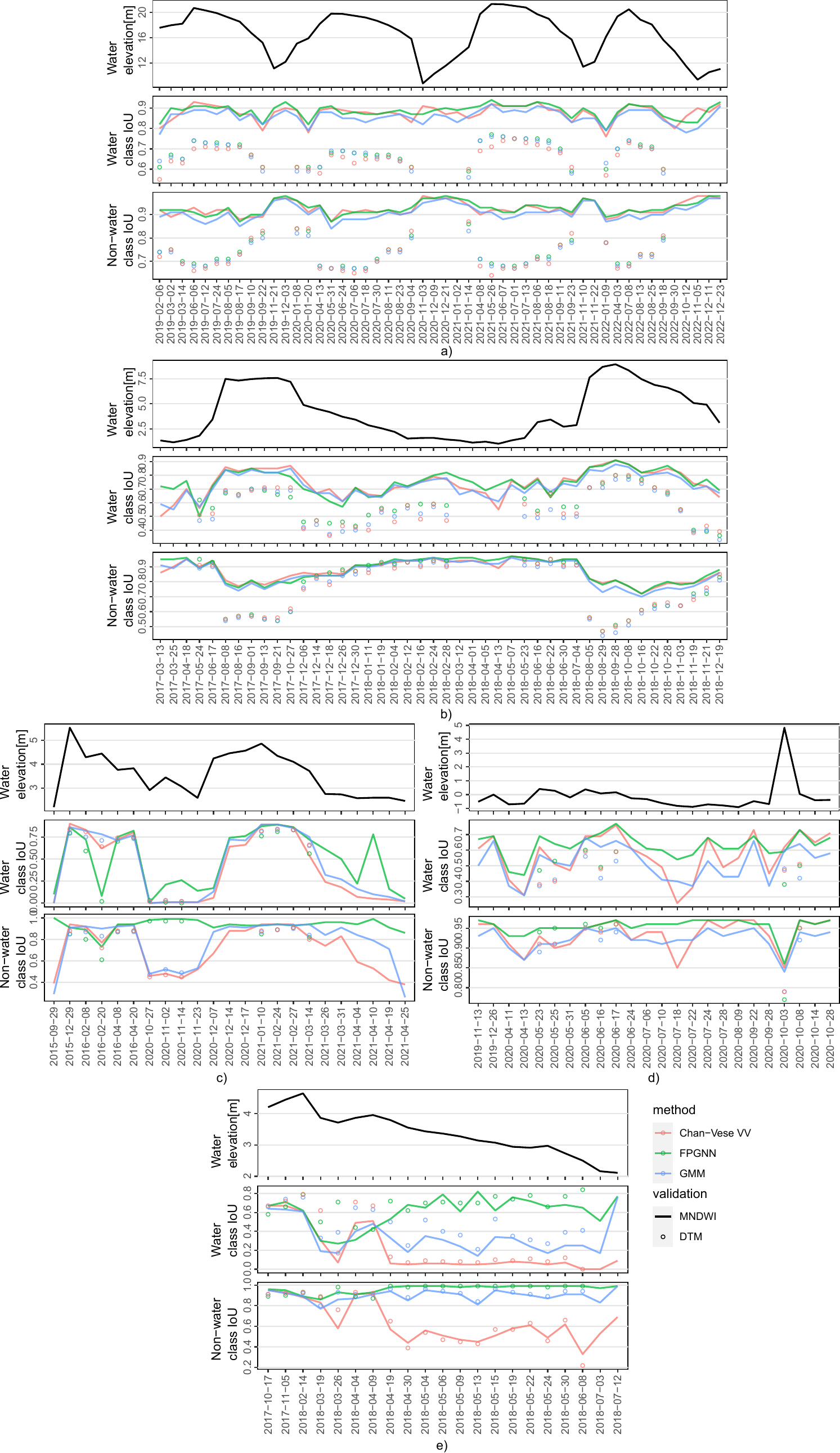}
\par\end{centering}
\caption{\label{Fig 4} Comparisons of IoU for days of optical image acquisition
for sites located in Brazil (a), Cambodia (b), Italy (c), England
(d) and Poland (e). Panel for each site includes three plots: observed
water elevation, calculated water and non-water class IoU for the
FPGNN, GMM and SC methods in MNDWI and DTM validation.}
\end{figure*}

For the study cases in Brazil and Cambodia (Fig. \ref{Fig 4}a, b),
there are some visible dependences between the IoU and the water level.
The IoU of the water class increased when the water level in the river
increased, and conversely for the non-water class, in the MNDWI validation.
This dependence is more visible for the DTM validation results due
to the greater variation of the IoU. The IoU of the water class increased
during floods in almost the whole analyzed area, so each misclassification
has a smaller impact on the IoU value than when the water level is
low during the dry season. A similar relationship can be noticed for
the Cambodia site. In this case, our method provided a higher IoU
for water class than the Chan-Vese VV and GMM algorithms during low
water elevation, i.e. when the difference in the water class IoU was
the greatest.

For the study area in the UK, the lowest IoU values were obtained
when low water elevation was in the river independently of the method
used (Fig. \ref{Fig 4}c). The imaged river is 20 m wide, which is
only double the pixel size of the SAR images. Therefore, the river
area when the water level is low is less visible on the SAR data,
while it is detected on the MNDWI. The benchmarking methods performed
better in that case and classified the river as water class; however,
at the same time, it misclassified many agricultural fields. This
resulted in a low IoU for both classes during low water elevation.
For the same cases, our algorithm could not detect the river area
but classified the non-water areas better in both validation methods.
On 20 February 2016, the water class IoU for the FPGNN was considerably
lower than for the Chan-Vese and GMM methods. The registered backscattering
for the water is higher than on other days due to waves on the water\textquoteright s
surface. The FPGNN mostly analyzed the texture of the image during
training. The higher intensity of the water pixels on the image from
20 February 2016 compared to the images from the other days caused
the water to not be detected by our method.

The results for the Italy area are characterized with steady IoU values
above 0.9 for the non-water class (Fig. \ref{Fig 4}d). The IoU of
the water class is significantly lower than that of the non-water
class because water occupies only a small part of the area. Most of
the water areas are located inside the river valley and appear outside
it only during floods. There is one exception, in which the IoU for
both classes decreased significantly (3 October 2020). This low result
was caused by the significant difference between the radar and MNDWI
images, which were obtained during a flood event. The optical image
was acquired about 7 hours earlier than the SAR image. The registered
flood was a dynamic phenomenon, which resulted in a considerable change
in the water extent between the acquisitions. Other days with a lower
water class IoU (below 0.5) are 11 and 13 April 2020, where farmlands
without vegetation were misclassified in the SAR data as water due
to their smooth surface. However, our method obtained a higher IoU
for both classes than the Otsu because it misclassified fewer agricultural
fields. This is the effect of extracting contextual information by
the convolutional layers in our model. 

The IoU for the area in Poland increased with the decreasing water
elevation for our method (Fig. \ref{Fig 4}e). The benchmarking algorithms
obtained similar results for the water class as our method, however,
their IoU decreased significantly when the water elevation in the
river started decreasing. This was caused by numerous misclassified
pixels of the non-water class by Otsu. The FPGNN performed better
independent of the water elevation, especially during the low water
stage, and obtained a significantly higher IoU for both classes. The
lowest IoU for the water class (below 0.4) was for images from the
days between 19 March and 9 April 2018 in the MNDWI validation (for
FPGNN and the benchmarking methods). This was caused by the optical
images being registered when the river and the floodplains were partially
or completely frozen. The ice on the river has higher backscattering
on radar images than open water \cite{Yan2020}. Effectively water
pixels were only labeled on the SAR data in places where the ice had
melted. 

\subsection{Visual interpretation of results}

\begin{figure*}
\begin{centering}
\includegraphics[width=0.9\textwidth]{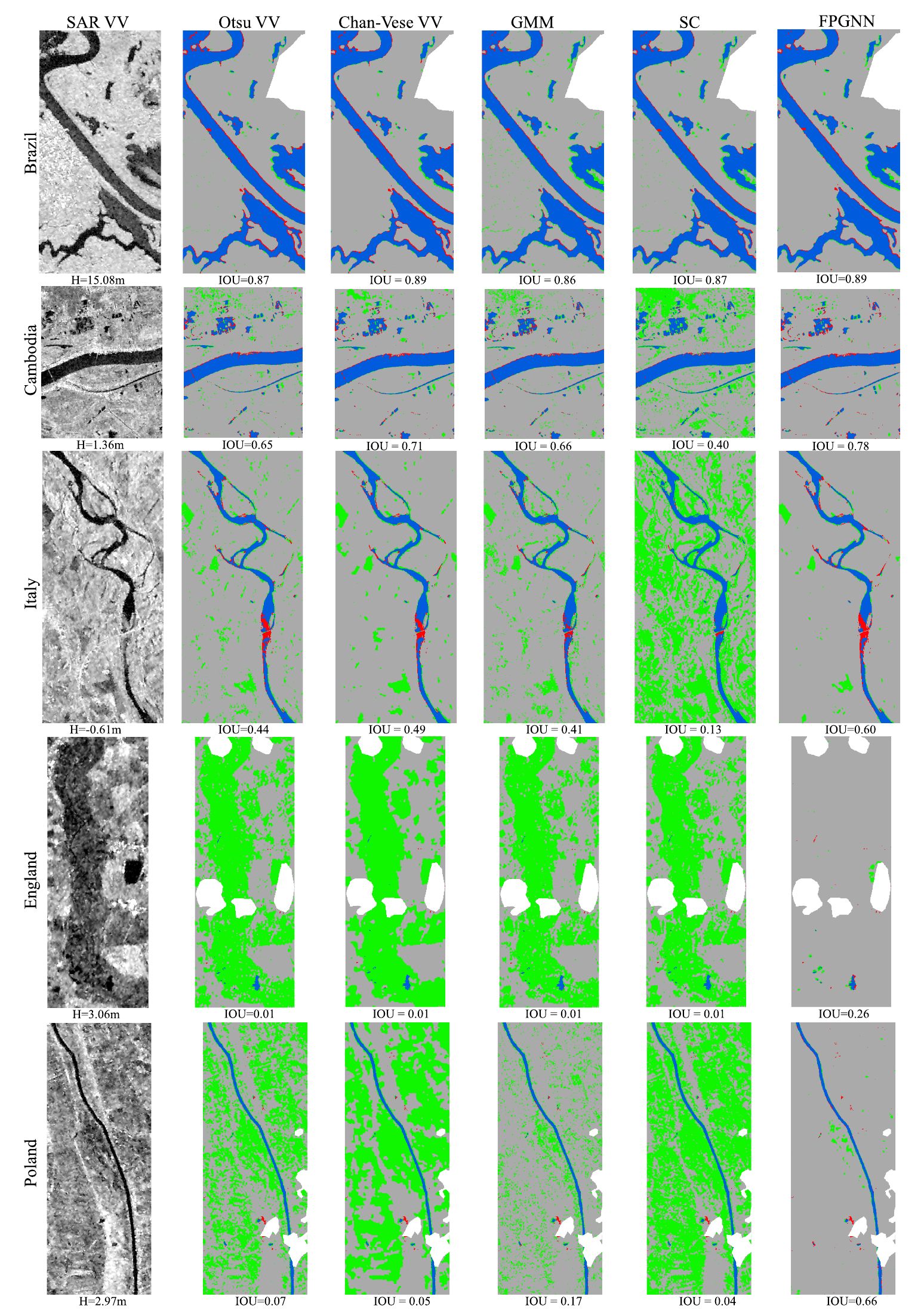}
\par\end{centering}
\caption{\label{Fig 5} Contingency maps obtained from a pixel-to-pixel comparison
between the MNDWI water masks and FPGNN or the benchmarking methods
for all study sites for low water elevation. True positives are shown
in blue, true negatives in gray, false positives in green, and false
negatives in red.}
\end{figure*}

\noindent 
\begin{figure*}
\begin{centering}
\includegraphics[width=0.9\textwidth]{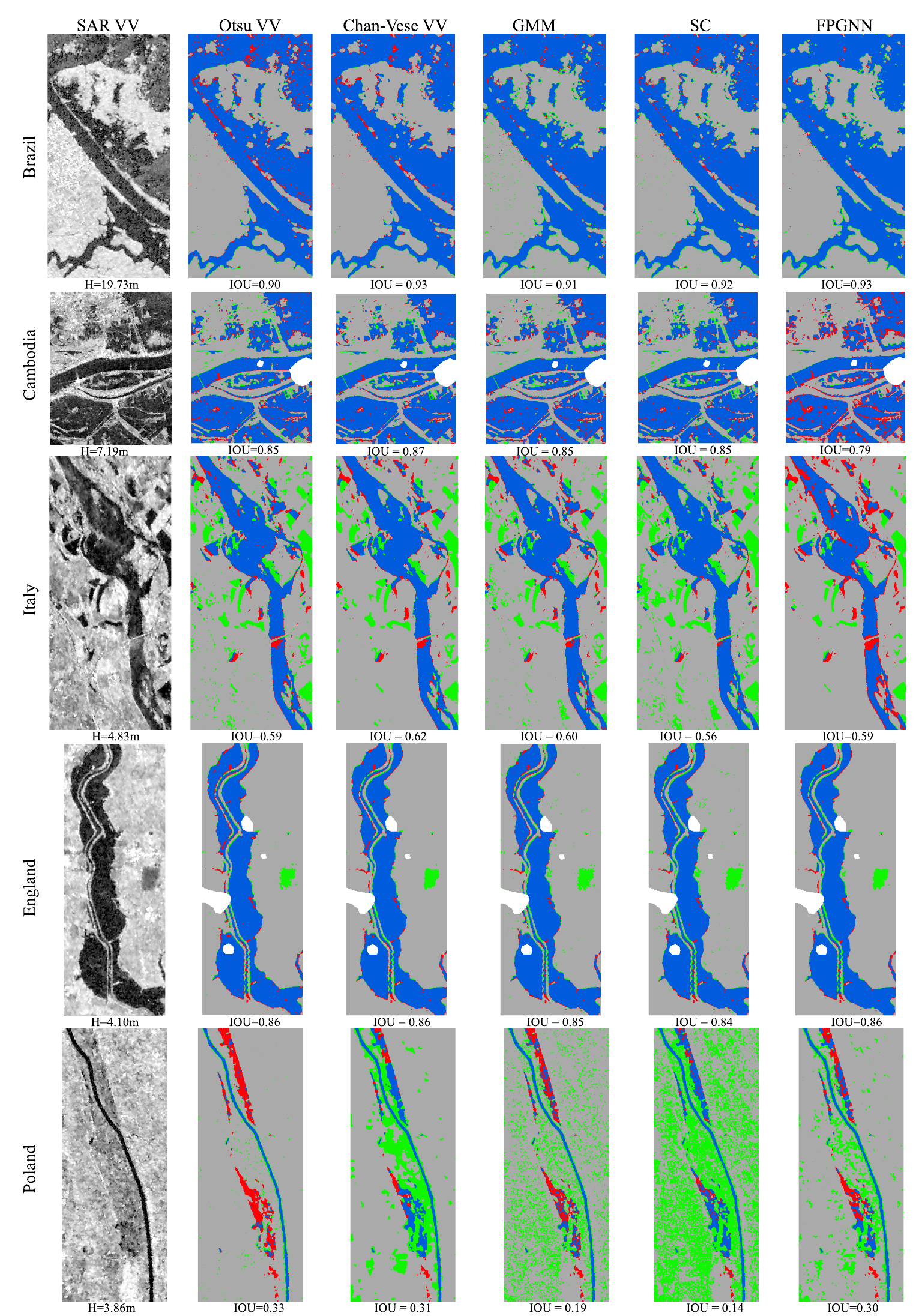}
\par\end{centering}
\caption{\label{Fig 6} Contingency maps obtained from a pixel-to-pixel comparison
between the MNDWI water masks and FPGNN or the benchmarking methods
for all study sites for high water elevation. True positives are shown
in blue, true negatives in gray, false positives in green, and false
negatives in red.}
\end{figure*}

Based on visual comparisons of the water masks obtained with data
from various sensors, we noticed the main drawbacks of radar and optical
satellite data (Fig. \ref{Fig 5}). The main visible difference in
the water masks from SAR and MNDWI from the optical sensor was the
shift in the water area. It was caused by the fact that the SAR sensor
sent its signal sideways, and the terrain was illuminated only from
one side. Tall objects along the river, such as trees and buildings,
also influenced the shift of the mapped water area in the SAR image.
Also, despite the speckle filtering in SAR, some noise in the pixels
remained. The noise decreased the classification efficiency, especially
for pixel-based classification method. In contrast, the water mask
obtained with optical data was more homogeneous. Another observed
effect on the SAR water masks was their worse preservation of the
narrow terrain objects than on the MNDWI images due to the speckle
filtering process. The water areas on the radar images were identified
by the low intensity of backscattering from their smooth surface.
The roughness of the water surface could change due to strong wind
or possible water cascades, which resulted in increasing the backscattering
and misclassification of pixels (the Italy area). In contrast, such
phenomena did not influence the water extent based on the optical
images. Moreover, the SAR sensor was more sensitive at detecting objects
on the water surface, such as bridges, due to their high roughness
compared to water. On the MNDWI results, these objects were not classified
as non-water because of the low spatial resolution of the short-infrared
band.

The differences between the results for Otsu and FPGNN were not only
in their IoU values but also in visual comparison. One of main advantages
of Otsu was that it better prevented narrow line objects, such as
tributary rivers, roads and bridges. The FPGNN omitted those objects
due to the applied 3x3 filters in the convolutional layers and the
2x2 windows in the polling layers. Other studies showed that supervised
deep learning methods could quite accurately detect line objects \cite{Sadiq2022}.
However, our method did not use any ground-truth water masks, hence,
it lost this information. On the other hand, this allowed the FPGNN
to classify correctly the single, bright pixels in the water segments,
which were misclassified by the Otsu thresholding. A similar effect
of more homogenous water segments achieved the Chan-Vese segmentation.
However, this method slightly enlarged the water segments, which negatively
influenced the water map accuracy. The FPGNN performed significantly
better than benchmarking methods in distinguishing between water bodies
and agricultural areas without vegetation because the neural network
extracted contextual features from the images. 

For the sites in Brazil and Cambodia, the water masks for FPGNN and
benchmarking methods were similar for SAR images registered during
low and high water elevation. In the case of the England, Poland,
and Italy study areas, the accuracy achieved for flood images was
comparable. However, when the water regions occupied a small part
of the areas analyzed, the performance of all of the tested benchmarking
methods considerably decreased, and there were visible lots of misclassified
pixels. The Otsu method has been known to provide unstable results
when there is a small area of water bodies and large non-water features
in the image \cite{Yang2017}. Whereas, the FPGNN correctly classified
the water segments on those specific images. Our method produced more
accurate results, especially for SAR images with no visible river
(the England area). 

\subsection{Advantages and limitations}

The main advantage of our method is that it does not require labeled
data showing water areas prepared by experts. The approach needs only
the preprocessed times series of radar data and the water elevations
for the corresponding days. The SAR data preprocessing steps can be
automated by remote sensing software or performed using online APIs
(ASF, Sentinel Hub). Hydrological observations do not need to be processed;
however, requiring these data limits the method\textquoteright s application
to river areas in which water gauge stations are present. However,
our study shows that the method works not only in the proximity of
the river gauge, but at least 7 km upstream or downstream. We expect
this distance should be greater, i.e. our method should be valid for
a regional-size catchment scale, which should be verified in a follow-up
study. 

The most important limitation is that the method does not work for
river areas where the imaged riverbed does not change its width while
the elevation of the water changes significantly. In that case, there
is no relation between the flooded area and the elevation of the water,
and the model cannot learn from those data. Another limitation is
the necessity to train the new model for each area. We expect that
the transfer learning approach should be beneficial in our method,
however, this should be verified in a follow-up study. In addition,
the input data should contain the SAR images and water elevation presenting
the river during different stages (flood, drought). Based on the England
area, we noticed another limitation -- the width of the river. Our
approach performed better for rivers at least twice as wide as the
spatial resolution of the input radar data, including when the river
stage is low. However, this drawback can be solved by applying images
with higher spatial resolution. The results of this study suggest
that the frequent presence of ice on the surface of the water could
significantly influence the action of our method due to the lower
correlation between water extent area and water elevation. A similar
problem could occur in areas with vegetation obscuring the water.
In addition, data such as optical satellite images or DTM are required
if one wishes to fully verify the results against ground-truth. 

\section{Summary}

In this paper, we proposed a physics-guided neural network for flood
area detection (FPGNN) using SAR data and river gauge observations.
The assumption in our approach was a monotonically non-decreasing
relation between a local river water elevation and the flooded area.
We applied time-series SAR images and water elevation observations
as input datasets in our method. The FPGNN was tested for five study
sites characterized by various climates and land cover types. 

Our results show that the relationship between SAR-derived flood extents
and water elevations occurs and can be used to train a neural network
by applying the correlation coefficient between the mentioned variables
as a loss function. The FPGNN provides classification models that
produce more accurate water masks for different areas than the compared
methods. The highest differences between results are for water maps
showing rivers during low water elevations, when small part of SAR
image is covered by the water. While, the other unsupervised approaches
mismark the water class in many pixels. However, the application of
FPGNN is limited by river valley characteristics, such as the shape
of the riverbed, the density and height of vegetation, and the occurrence
of ice cover, which can decrease the correlation between the water
extent and the water elevation. 

\bibliographystyle{IEEEtran}
\addcontentsline{toc}{section}{\refname}\bibliography{references}

\end{document}